\documentclass[11pt, a4paper, copyright]{techreport}

\usepackage[square, numbers]{natbib}
\usepackage{subcaption}
\usepackage{multirow}
\usepackage{adjustbox}
\usepackage{multicol}
\usepackage{xspace}
\usepackage[bottom]{footmisc}
\usepackage{subfiles}

\addto\extrasenglish{
}

\usepackage{amsmath}
\usepackage{booktabs}
\usepackage{enumitem}

\renewcommand{\phi}{\varphi}

\renewcommand{\epsilon}{\varepsilon}
\renewcommand{\imath}{\mathrm{i}}

\newlength{\restsubwidth}
\newlength{\restsubheight}
\newlength{\restsubmoreheight}
\newcommand{\rest}[2]{%
        \settowidth{\restsubwidth}{\ensuremath{#2}}
        \settoheight{\restsubheight}{\ensuremath{{}_{#2}}}
        \ensuremath{{#1\hskip 0.5pt}_{\vrule\kern2pt\parbox[b][%
        4pt][b]{\the\restsubwidth}{%
                        \ensuremath{{}_{#2}}}}}
        }

\newcommand{\sysname}{\textit{MEgoVista}\xspace}
\newcommand{\depname}{\textit{MEgoDeploy}\xspace}
\newcommand{\device}{\textit{MEgo View}\xspace}

\newcommand{\stageb}{\textsc{HandStage}\xspace}

\newcommand{\Lrom}{\ensuremath{\mathcal{L}_\mathrm{rom}}\xspace}

\newcommand{\Lbend}{\ensuremath{\mathcal{L}_\mathrm{bend}}\xspace}
\newcommand{\Lomega}{\ensuremath{\mathcal{L}_\omega}\xspace}
\newcommand{\Lpose}{\ensuremath{\mathcal{L}_\mathrm{pose}}\xspace}
\newcommand{\Lreproj}{\ensuremath{\mathcal{L}_\mathrm{reproj}}\xspace}
\newcommand{\Lwrootvel}{\ensuremath{\mathcal{L}_\mathrm{wrootvel}}\xspace}
\newcommand{\Lwrootang}{\ensuremath{\mathcal{L}_\mathrm{wrootang}}\xspace}
\newcommand{\Lfingervel}{\ensuremath{\mathcal{L}_\mathrm{fingervel}}\xspace}
\newcommand{\Lsdepth}{\ensuremath{\mathcal{L}_\mathrm{sdepth}}\xspace}

\newcommand{\Lsmooth}{\ensuremath{\mathcal{L}_\mathrm{smooth}}\xspace}

\newcommand{\chingmu}{Chingmu\xspace}

\newcommand{\pampjpe}{PA-MPJPE-p\xspace}

\newcommand{\ctp}{CT-p\xspace}
\newcommand{\ate}{ATE\xspace}

\titlespacing*{\section}   {0pt}{1.7ex plus .3ex}{0.8ex}
\titlespacing*{\subsection}{0pt}{1.3ex plus .2ex}{0.6ex}
\institution{Maniformer}
\reportseries{Technical Report}
\reportnumber{001}

\title{\centering \sysname: Multi-view Ego-aware Motion Estimation for Metric 4D Hands and Head in the Wild}

\author{{\fontsize{14}{18}\selectfont\bfseries Maniformer.ai}}

\begin{abstract}
Learning manipulation from human video requires high-fidelity hand-motion reconstruction in metric units.
Today's metric hand labels come from studio rigs and instrumented headsets, and both are confined
in the same two ways: neither leaves a prepared setting, and neither is checked against an
independent reference. Unconstrained head-worn recording promises the opposite trade-off, scaling
with the number of people wearing a device. We therefore introduce \sysname, an offline pipeline
that turns a single unprepared \device recording into metric two-hand and head motion in one
gravity-aligned world frame. Three properties set it apart from existing egocentric reconstruction
systems: first, it reconstructs in settings studio volumes and tabletop rigs cannot reach,
settling hand ownership at detection so bystander hands stay out of the wearer's trajectory;
second, it takes its metric gauge from calibrated stereo rather than a monocular prior, installing
scale at initialisation so policies receive physical units, not arbitrary coordinates; third, both
outputs are scored inside a motion-capture volume against independent \chingmu optical capture,
under a protocol that audits its own reference and charges what a method declines to predict.
\sysname is offered as a measured route from egocentric video to metric hand supervision, one that
widens where such labels can be gathered.To bridge the gap between lab validation and industrial-scale data annotation, we further introduce \depname, a production-grade GPU stack that fully preserves reconstruction fidelity while achieving a 14.05× end-to-end throughput speedup, scaling the generation of metric 4D pseudo-labels for embodied foundation models.
\end{abstract}

\begin{document}
\maketitle

\section{Introduction}
\label{sec:intro}

Egocentric video has become a central data source for embodied AI. Learning a manipulation policy
requires observations paired with the physical states and actions that produced them, but collecting
such pairs on a real robot is expensive: every hour of teleoperation consumes hardware, an operator,
and an environment reset~\citep{datapyramid2026}. Human recording offers a different trade-off. A
head-mounted camera captures real contact physics, dexterous five-fingered hands, and the diversity
of everyday environments, and it requires no robot in the collection loop, so the collection rate
scales with the number of people wearing a device rather than the number of available
robots~\citep{ego4d,epickitchens}. Egocentric datasets have grown accordingly, from hundreds of
hours of passive activity capture to recent releases measured in thousands of
hours~\citep{egoexo4d,egodex,egocentric10k}, and egocentric data is now a standard component of the
pretraining mixtures used by embodied foundation models~\citep{datapyramid2026}. Within such data,
hand motion is the signal that matters most. Semantic annotations such as narrations and verb-noun
labels describe what a person did, but not the motion that accomplished it. Every method that
converts a human demonstration into a robot action space operates on the hand: the wrist trajectory
is mapped through inverse kinematics, and finger articulation, typically parameterised by
MANO~\citep{mano2017}, is transferred by morphology-aware
retargeting~\citep{ph2d,egoscale2026,vitra2025}. The quality of the resulting supervision is
therefore bounded by the accuracy of the estimated metric 3D hand pose.

\noindent Despite growing adoption, egocentric collections that carry metric 3D hand labels remain narrow.
Studio rigs such as ARCTIC and Assembly101 deliver accurate poses under multi-view optical motion
capture, but cannot leave their capture volume~\citep{arctic,assembly101}. Instrumented headsets
extend that range through device-native inside-out tracking: EgoDex recorded 820 hours of bimanual
manipulation using Apple Vision Pro~\citep{egodex}, but remains confined to a prepared tabletop.
Large-scale passive corpora such as Ego4D capture in-the-wild activity but ship no metric 3D
supervision~\citep{ego4d}. Where 3D labels are released, their accuracy is asserted from hardware
specifications rather than established against an independent metric reference.

\noindent Recent work has advanced hand reconstruction through learned regressors and temporal optimisation.
WiLoR and HaMeR predict MANO parameters from a single crop~\citep{wilor,hamer2024}, while HaWoR and
Dyn-HaMR lift monocular sequences into world-frame trajectories~\citep{hawor2025,dynhamr2025}. These
methods obtain metric depth from monocular learned predictors such as UniDepthV2~\citep{unidepthv2},
whose scale is a prior rather than a measurement, and the reprojection term cannot correct depth
error along the viewing ray. Standard detectors distinguish only left from right, not wearer from
bystander, corrupting both precision and recall when other people's hands appear. Occlusion remains
problematic: the object, forearm, and frame edge hide the hand during the contact phases that matter
most.

\noindent Head motion tracking faces a related challenge. Head-worn capture is rotation-dominated, so
translation over a keyframe interval is often negligible. Visual-inertial trackers such as ORB-SLAM3
recover metric scale from the inertial unit~\citep{orbslam3}, but require a usable baseline between
successive keyframes, and when rotation dominates, scale estimation degrades. Offline
structure-from-motion pipelines such as COLMAP reconstruct more accurately~\citep{colmap2016}, but
cannot observe scale monocularly and produce trajectories whose metric scale is arbitrary.

\noindent To address these challenges, we present \sysname, a system that reconstructs metric 3D hand and head
motion from egocentric video recorded in unconstrained real-world environments. Unlike prior methods
that remain confined to prepared tabletops or studio volumes, our approach operates on footage
captured during everyday occupational tasks. The system resolves multi-person ambiguity, recovers
metric depth without monocular learned priors, handles occlusion across multiple manipulation phases,
and maintains stable head tracking when the wearer turns or moves. We validate the reconstruction
accuracy inside a motion-capture volume and report millimetre-scale error against an independent
metric reference. Our contributions are:

\begin{itemize}[leftmargin=*,topsep=2pt,itemsep=1pt]
\item \textbf{Robust reconstruction in unconstrained real-world settings.} The system handles
      in-the-wild occupational environments where studio rigs and tabletop setups cannot operate. It
      addresses multi-person/multi-hand ambiguity, occlusion robustness, and rotation-dominated head
      motion, delivering supervision in settings that existing methods do not cover.
\item \textbf{Superior capture specifications and reconstruction accuracy.} Our capture achieves a
      wider field of view than prior egocentric systems, and reconstruction accuracy reaches
      sub-centimetre scale: 0.65 to 2.83\,mm for head trajectory and 4.29\,mm for finger articulation,
      measured against an independent motion-capture reference.
\item \textbf{Trustworthy metric scale and coordinates.} Our system recovers
      reliable metric scale and gravity-aligned coordinates, ensuring that downstream
      manipulation policies receive hand trajectories in physical units rather than arbitrary coordinates.
\end{itemize}

\section{Related Work}
\label{sec:related}
\label{subsec:related-datasets}
\label{subsec:related-hands}
\label{subsec:related-endtoend}
\label{subsec:related-sfm}
\label{subsec:related-protocol}

\textbf{Egocentric hand reconstruction.} MANO makes hand recovery a fitting problem~\citep{mano2017}, and
single-crop regressors predict its parameters with no metric depth and no coupling between
frames~\citep{hamer2024,wilor} --- our initialiser, not a competitor. HaWoR, Dyn-HaMR and HaPTIC lift
monocular hand motion into a world frame~\citep{hawor2025,dynhamr2025,haptic2025} and produce the
output shape we do; they estimate that frame from the same stream that supplies the hand
evidence, whereas we take it from calibrated hardware. MS-MANO routes pose through a muscle-tendon
simulator that cannot represent an impossible configuration~\citep{msmano,physcap}, and BioPR learns
the equivalent priors~\citep{biopr}; we approximate the effect with differentiable barriers.
Feed-forward geometry collapses the stages~\citep{dust3r,mast3r,vggt} and removes the point at which
a known baseline can be injected. Learned depth, stereo and segmentation supply per-frame
geometry~\citep{unidepthv2,unik3d,foundationstereo,sam2}; we consume these as initialisation and
gate rather than trust them.

\noindent\textbf{Stereo depth and multi-view geometry.} Monocular depth estimation from learned predictors
such as UniDepthV2~\citep{unidepthv2} and UniK3D~\citep{unik3d} provides per-frame depth but produces
scale as a prior rather than a measurement, and the scale varies across frames. Stereo methods recover
metric depth from calibrated baselines. Classical approaches such as block matching and
semi-global matching require rectification, which distorts wide-angle fisheye views at the image
periphery where egocentric hands appear. Recent learned stereo methods such as
FoundationStereo~\citep{foundationstereo} operate on unrectified views and generalise across camera
models, making them suitable for the high-vergence pairs formed by a fisheye camera and a lateral
camera in a head-mounted rig. Multi-view geometry frameworks such as COLMAP~\citep{colmap2016} and
generalised camera models~\citep{generalizedrelativepose,colmaprig} treat rigidly mounted cameras as
one sensor with known relative poses, reducing drift and enabling scale recovery from a fixed
baseline. We adopt this constraint and additionally use the calibrated baseline as the gauge that
fixes metric scale across the entire reconstruction.

\noindent\textbf{Motion Tracking.} Head-worn capture is rotation-dominated, often with negligible
translation over a keyframe interval, and wide-angle; the useful sort of this literature is by what
supplies metric scale. ORB-SLAM3 makes scale observable from acceleration~\citep{orbslam3} and is the
online VIO pipeline we originally built \device's head-pose stage around, so it is at once a
precursor of our system and our comparison point for the offline path --- refined offline with a
full-sequence bundle adjustment and per-camera-frame re-registration for that comparison, rather than
scored as its raw real-time trajectory. Offline pipelines reconstruct more accurately than an online
tracker can~\citep{colmap2016,glomap,megasam2025} but cannot observe scale monocularly, and
learned-prior loops~\citep{droidslam,mast3rslam} could in principle carry a fixed-scale edge as ours
does --- DROID-SLAM is our second baseline in \autoref{tab:head-gt}. Treating rigidly mounted cameras
as one generalised camera with known relative poses~\citep{colmaprig,generalizedrelativepose} under a
fisheye projection~\citep{kannalabrandt2006} is our head stage's machinery; the literature adopts
that constraint to reduce drift, and we additionally use the calibrated baseline as the \emph{gauge}.

\section{\sysname}
\label{sec:method}
\label{sec:platform}
\label{subsec:rig}
\label{subsec:calibration}
\label{subsec:delivery}
\label{subsec:pipeline}
\label{subsec:implementation}

\subsection{Hardware}
\label{subsec:handfront}
\label{subsec:stereodepth}
\label{subsec:handopt}
\label{subsec:priors}
\label{subsec:qc}
All data is collected using \device\footnote{\url{https://www.maniformer.ai/en/mego}}, a
custom-designed head-mounted device for acquiring human manipulation behaviour in the wild.
\device carries fisheye
cameras~\citep{kannalabrandt2006} whose wide field of view is analogous to human binocular vision,
three of which are used below: a central camera and a calibrated stereo pair. The high resolution
and 60\,fps frame rate facilitate fine-grained and agile hand motion tracking, while the integrated
Inertial Measurement Unit (IMU), sampled at 500\,Hz, enables improved accuracy in camera pose
estimation. 

\begin{figure*}[t]
\centering
\includegraphics[width=1.0\textwidth]{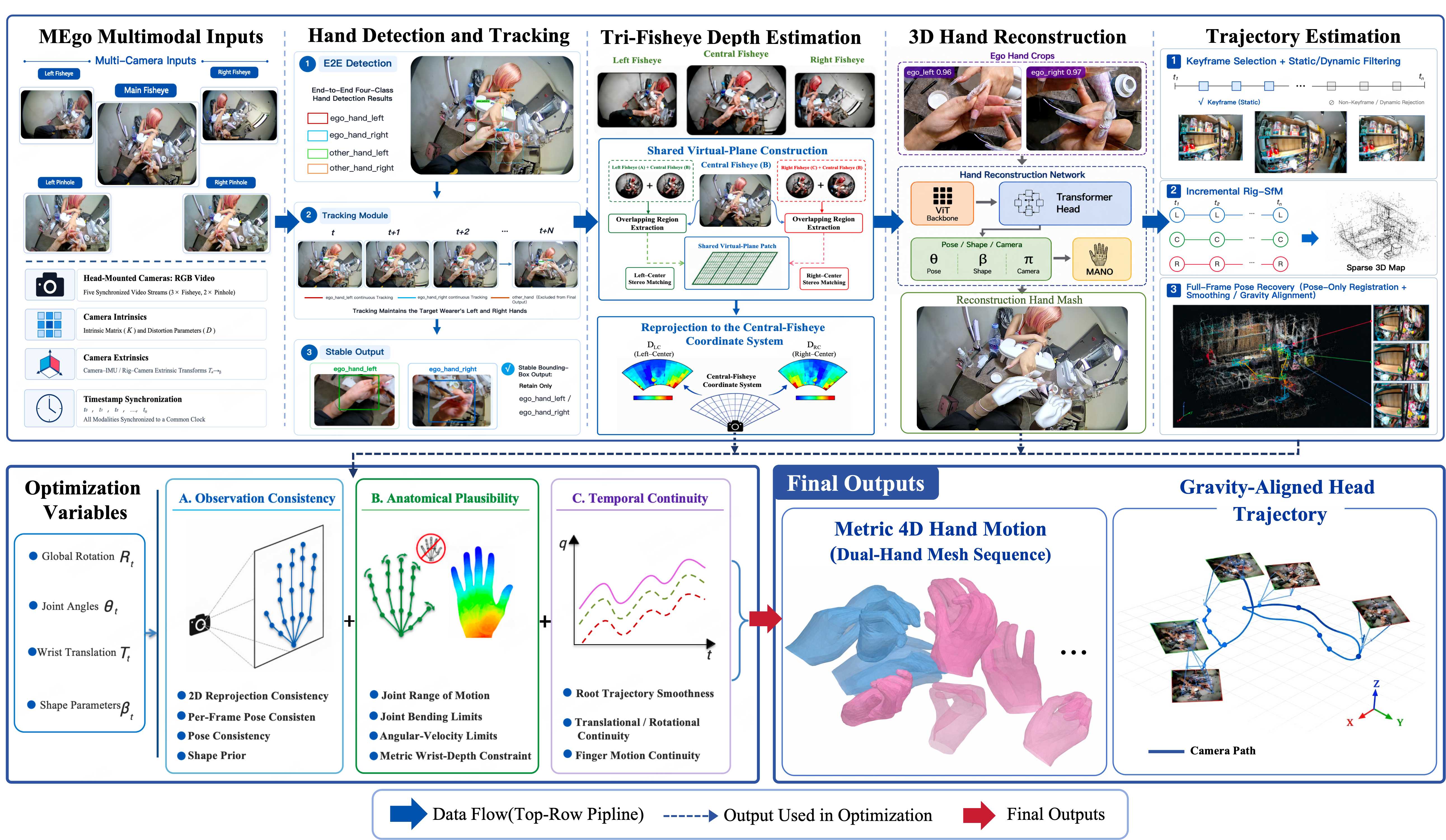}
\caption{
\textbf{The \sysname pipeline.} A raw recording enters at the left. \sysname --- generate the per-frame camera pose.
\stageb --- hand
detection, MANO regression, multi-view depth estimation, full-batch bundle adjustment. Outputs at the right: head trajectory and hand trajectory.
}
\label{fig:pipeline}
\end{figure*}

\subsection{Hand Reconstruction}
\label{subsec:handfront}
\label{subsec:stereodepth}
\label{subsec:handopt}
\label{subsec:priors}
\label{subsec:qc}

\par\noindent\textbf{A. Ego-Aware Hand Detection and Initialization.}
In-the-wild egocentric videos frequently contain multiple visible hands from both the camera wearer
and surrounding people. We therefore employ a trained end-to-end detector that jointly estimates
hand location, ownership, and handedness, allowing the wearer's left and right hands to be directly
distinguished from other visible hands. A lightweight tracker further maintains temporally stable
hand identities and consistent bounding boxes across the sequence. The tracked wearer-hand regions
are then processed by models akin to WiLoR~\citep{wilor} and HaMeR~\citep{hamer2024} to obtain
per-frame MANO~\citep{mano2017} pose and shape estimates in the camera coordinate system, which
serve as initialization for the subsequent metric reconstruction and sequence-level optimization.
\par\noindent\textbf{B. Tri-Fisheye Virtual-Plane Metric Depth Estimation.} 
To address the challenges posed by 
fisheye cameras on head-mounted devices---namely their large field of view, wide baselines, and mounting
pose deviations---this paper proposes a stereo depth estimation and
fusion method based on a triple-fisheye camera system. The central
camera (MID) on the device forms two wide-baseline stereo pairs,
MID$\times$LEFT and MID$\times$RIGHT, with the two side cameras (LEFT
and RIGHT). For each pair, we design a wide-baseline common-rotation
rectification scheme: the rectified coordinate frame takes the baseline
direction as its $e_1$ axis to absorb arbitrary three-axis mounting
offsets between the cameras, and the angle bisector of the two optical
axes as its $e_3$ axis, so that the common field of view is centered in
the rectified images. This design guarantees row alignment while
maximally preserving the effective field of view. The two rectified
stereo pairs are then processed by the FoundationStereo~\citep{foundationstereo} network, each
producing a dense depth map together with a per-pixel confidence measure
derived from the matching probability volume. The two depth maps are
arbitrated and fused within the common field of view of the MID view.
Finally, the fused depth is back-projected and analytically reprojected
onto the MID fisheye pixel grid via the KB fisheye model, yielding a
dense depth map that is pixel-aligned with the original fisheye image.
\par\noindent\textbf{C. Full-Sequence Hand Motion Optimization.}
A per-frame regressor produces hands that may be individually plausible but are not necessarily
consistent over time: they can jitter, drift in depth, or pass through configurations that a real
hand cannot reach. We therefore optimize a single MANO trajectory over the complete episode,
subject to three complementary requirements: consistency with the observations, continuity of the
3D motion, and anatomical plausibility.

\noindent Before optimization, we select reliable and informative observations to prevent erroneous
per-frame estimates from contaminating the full-sequence trajectory. Each frame is evaluated
according to detection confidence, depth validity and dispersion, joint spread, bone-ratio
consistency, and left--right agreement. Observations that fail these checks are excluded from the
corresponding optimization terms, while the remaining high-quality observations are used to
constrain the full-sequence optimization.

\noindent The objective is organized accordingly:

\begin{enumerate}[leftmargin=*,topsep=1pt,itemsep=0pt]
\item \textbf{Tight to the observations.}
      \Lreproj is the 2D keypoint residual, normalized by hand bounding-box area so a distant hand
      retains a meaningful contribution rather than being dominated by a nearby one;
      \Lsdepth constrains wrist camera-frame depth using the metric depth estimate from TriVP,
      providing the depth constraint that reprojection alone cannot reliably recover; and
      \Lpose anchors finger pose to the per-frame regressor.

\item \textbf{Continuous in 3D.}
      Second-order smoothness \Lsmooth acts on the root trajectory, while Huber terms
      \Lwrootvel and \Lwrootang suppress translational and rotational drift without allowing a
      single large motion to dominate the residual. \Lfingervel further suppresses single-frame
      articulation jitter.

\item \textbf{Anatomically correct.}
      Differentiable $C^1$ \texttt{relu}$^2$ barriers approximate, as soft penalties on a standard
      MANO solve, the anatomical constraints modeled explicitly by MS-MANO~\citep{msmano} and
      BioPR~\citep{biopr}. Specifically, \Lrom constrains the range of motion of individual joints,
      \Lbend enforces bending limits in output space rather than parameter space, and
      \Lomega limits angular velocity according to physiologically plausible bounds.
\end{enumerate}
Both hands are jointly optimized over the complete episode rather than independently frame by frame,
yielding temporally coherent MANO trajectories in the reconstructed metric world frame.

\subsection{Motion Tracking}
\label{subsec:vio}
\label{subsec:rigsfm}
\label{subsec:gauge}
\label{subsec:gravity}

\sysname recovers a head pose at very nearly every frame, and with it the metric, gravity-aligned
world frame the hand stage is solved in (\autoref{subsec:handfront}). A head pivots far more than
it travels, and no scene carries the metre or the vertical, so each quantity comes from the
instrument that determines it.
\par\noindent\textbf{A. Rig-Constrained Incremental Reconstruction.} Translation this weak is poor
evidence about the camera model: a bundle adjustment free to move it trades a trusted calibration
for lower reprojection error. The array is therefore one generalised
camera~\citep{colmaprig,generalizedrelativepose}, images sharing a decode index forming one rig
frame with a single head pose, all three fisheye under one Kannala--Brandt
model~\citep{kannalabrandt2006} whose calibration stays frozen. Selection divides the labour:
inertia finds motion cheaply, so gyroscope rotation and accelerometer energy nominate keyframes and
no fast turn is missed, while only vision knows whether a candidate is informative, so
Lucas--Kanade tracking against the last accepted keyframe rejects the poorly tracked and the
static. Visibility is declared from the same rig layout, not left to a sequential matcher blind to
time and camera arrays: keyframes match forward within each camera, the three at every shared
timestamp, and non-keyframes match adjacent mapped keyframes for later PnP. Mapping and
localisation are likewise split, a map dense enough for every frame being too large to
bundle-adjust: keyframes alone enter the incremental mapper~\citep{colmap2016}, whose finished map
is frozen and localised against by pose-only PnP.
\par\noindent\textbf{B. Calibrated-Stereo Metric Initialisation.} Scale can only be imposed at
initialisation, and neither obvious seed carries it: the automatic choice is a near-pure-rotation
pair whose scale can be wrong by orders of magnitude, and same-instant left and right images are
one rig frame, degenerate in generalised relative pose. The seed is therefore constructed:
left--right matches at a key timestamp, triangulated by DLT on the calibrated extrinsics and
refined by Levenberg--Marquardt, give one registered rig frame at the true baseline to continue
from. Scale thus comes from calibration rather than alignment --- though only in the seed, leaving
what grows from it free to drift.
\par\noindent\textbf{C. Post-Hoc Inertial Gravity Alignment.} A visual reconstruction has no preferred
vertical, and supplying one by joint visual-inertial bundle adjustment would let inertial drift
reach the geometry. Inertia therefore stays outside the optimisation, read only after vision is
complete: gravity is propagated through the 500\,Hz inertial stream, accelerometer contributions
down-weighted by $|\lVert a\rVert-g|$ so that violent motion is trusted less, smoothed by an RTS
pass, and applied as one global rotation into the initial reading's gravity-aligned frame.

\section{Experiments}
\label{sec:experiments}

The order of what follows is the argument. \autoref{sec:protocol} states the setup and the protocol
in full; \autoref{subsec:gt-audit} then measures how good the reference itself is, before anything is
measured against it; \autoref{sec:accuracy} and \autoref{sec:handacc} score the two delivered
quantities against that reference.

\subsection{Experimental Setup}
\label{sec:protocol}
\label{subsec:mocap}
\label{subsec:time-align}
\label{subsec:metrics}

\textbf{Dataset.} Every episode was recorded with \device at 60\,fps inside a 26-camera, 120\,Hz \chingmu
motion-capture volume, with the rig worn normally, uninstrumented, and
PTP-synced to the reference. The reference gives 6-DoF head pose from a
rigid marker cluster on the headset shell and a hand skeleton from a
15-marker-per-hand vendor solve.

\noindent\textbf{Metrics and alignment conventions.} Because alignment is part of
a metric's definition, not an afterthought, \autoref{tab:metrics} gives
each metric with its convention: every head figure here uses hand-eye
alignment with scale fixed at 1, so none is comparable to an \ate{} from
\texttt{evo}'s default similarity fit.

\begin{table}[htbp]
\centering
\caption{
\textbf{Metrics, each with the alignment that defines it.} Jitter and violation rate, are reference-free and unaligned. Fitted scale is reported beside every
trajectory figure but is not an error metric. \emph{Bold} marks the primary hand metric, not a best
result.
}
\label{tab:metrics}
\begin{adjustbox}{max width=\linewidth}
\begin{tabular}{llp{0.44\textwidth}}
\toprule
\textbf{Metric} & \textbf{Aligned by} & \textbf{What it measures} \\
\midrule
\textbf{\pampjpe} & Procrustes similarity: $R$, $t$, $s$ & \textbf{Primary.} Finger articulation alone, with hand scale and wrist orientation removed \\
\ctp     & none                    & absolute camera-frame wrist position; the depth-sensitive metric \\
\midrule
\ate RMSE     & hand-eye, scale fixed at 1 & per-frame head position residual; RPE@1\,s is the same residual over a 1\,s interval \\
Fitted scale  & similarity & metric gauge; not an error metric \\
\bottomrule
\end{tabular}
\end{adjustbox}
\end{table}

\subsection{Ground-Truth Construction and Validation}
\label{subsec:gt-audit}
\label{subsec:board}

\textbf{Spatial calibration puts both systems in one metric frame.} From paired
relative motions
$A_i$ and $B_i$ of the rig camera and the head-mounted marker cluster, we solve
$A_iX_{\mathrm{hc}}=X_{\mathrm{hc}}B_i$ for the
constant cluster-to-camera transform $X_{\mathrm{hc}}$, with intrinsics fixed.
At time $t$, the
measured cluster pose and $X_{\mathrm{hc}}$ carry each motion-capture point
into the camera frame.
Calibration captures are disjoint from validation and evaluation, so
$X_{\mathrm{hc}}$ is never
fitted to the hand predictions it is used to score.

\newlength{\gtpanelheight}

\begin{figure*}[t]
\centering
\captionsetup{justification=raggedright,singlelinecheck=false}
\captionsetup[subfigure]{justification=centering,singlelinecheck=false}
\setlength{\gtpanelheight}{0.325\textwidth}

\begin{subfigure}[t]{0.495\textwidth}
  \centering
  \includegraphics[
      height=\gtpanelheight,
      width=\linewidth,
      keepaspectratio
  ]{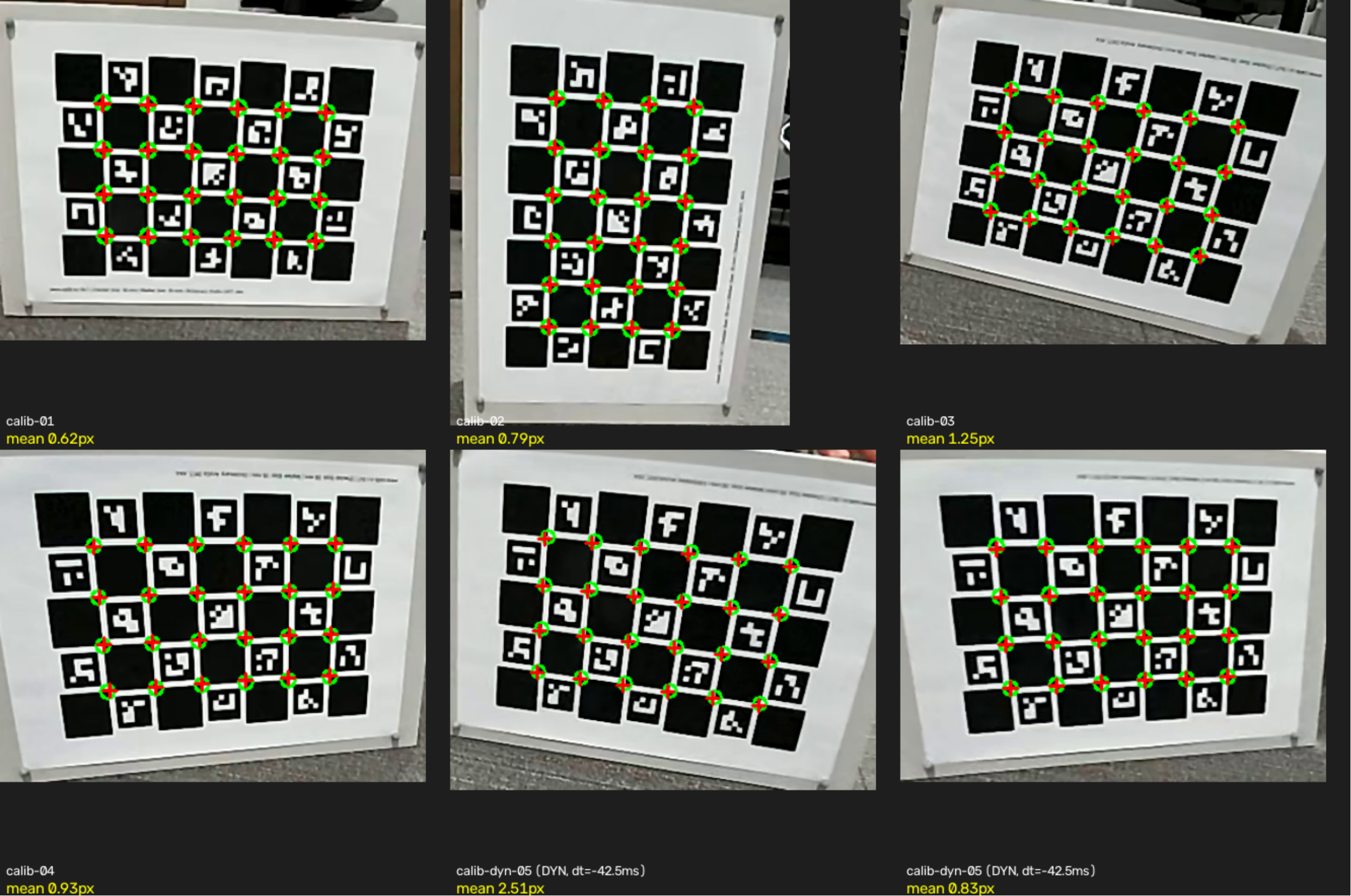}
  \caption{Held-out board poses.}
\end{subfigure}
\hfill
\begin{subfigure}[t]{0.495\textwidth}
  \centering
  \includegraphics[
      height=\gtpanelheight,
      width=\linewidth,
      keepaspectratio
  ]{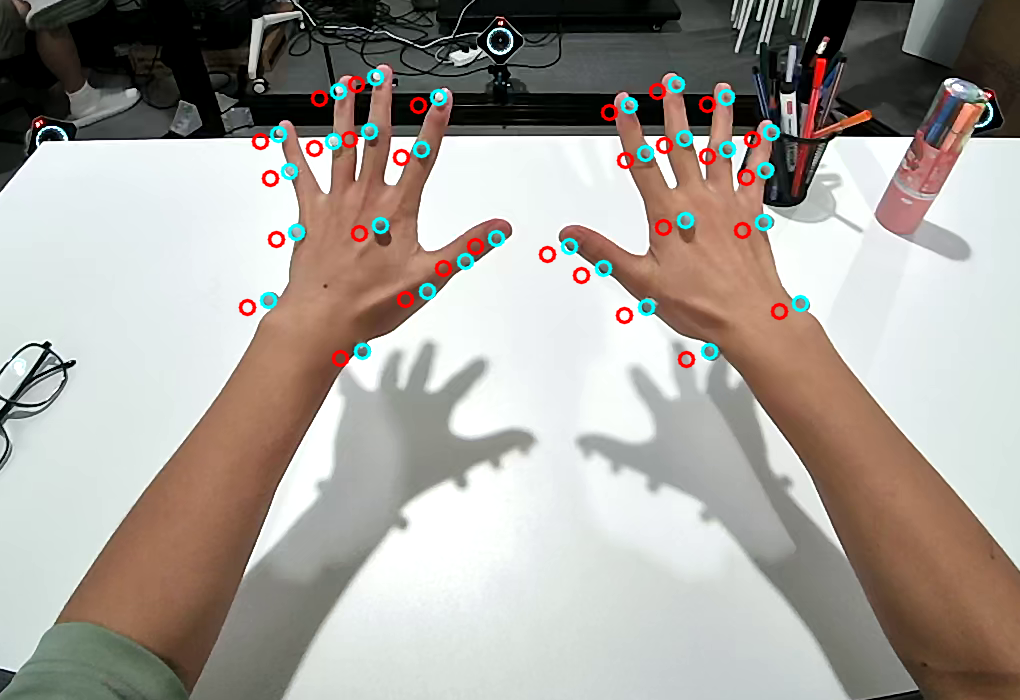}
  \caption{Hand points before (red) and after (cyan) calibration.}
\end{subfigure}

\caption{\textbf{Validation of the motion-capture-to-camera transform.}
  Left: projected corners on held-out static and moving board frames;
  annotations are per-frame means.
  Right: projected hand points before (red) and after (cyan) calibration.}
\label{fig:gt-validation}
\end{figure*}

\noindent\textbf{Held-out reprojection validates the chain end to end.} The board is
tracked by motion capture
and imaged simultaneously by the rig, then transformed and projected exactly as
the hand reference
will be. Representative frames in \autoref{fig:gt-validation} show mean errors
of 0.62--1.25\,px
when static and 0.83--2.51\,px in motion. Static frames test geometry; moving
frames additionally
exercise PTP timestamp correspondence. These are illustrative single-frame
means, not sequence-level
statistics. The hand overlay confirms that the calibration transfers to the
task domain.

\noindent\textbf{The remaining limitation is the solved hand reference.} Board
reprojection validates the
rigid transform, but not the vendor's marker-to-skeleton solver. The shared
screen of
\autoref{sec:protocol} therefore rejects observed failure signatures: frozen
fingers below
0.02\,mm/frame relative to the wrist (0.2--1.4 normally), solver spikes above
$6\times$ baseline,
and fingertip--marker distances above 40\,mm (10--20 normally). It catches
discrete failures but not
slow solver drift; any surviving reference error remains charged to the
evaluated pipeline.

\subsection{Motion-Tracking Accuracy}
\label{sec:accuracy}
\label{subsec:head-gt}
\label{subsec:head-diag}

\noindent As shown in \autoref{tab:head-gt}, ORB-SLAM3 and \sysname
  track the \chingmu reference to within a few millimetres and
  approximately $0.2^\circ$ across both capture sessions, without tracking
  loss. Neither method consistently dominates: \sysname is more accurate
  on Session 2 (0.65 versus 0.92\,mm steady-state), whereas ORB-SLAM3 is
  more accurate on Session 1 (2.12 versus 2.83\,mm). Removing the first
  5\,s of each episode reduces ORB-SLAM3's error by 1.15\,mm, compared
  with only 0.04\,mm for \sysname. This difference reflects the inertial
  initialisation required by ORB-SLAM3 but absent from our offline
  reconstruction. It motivates the use of the offline stage in production,
  where episodes last approximately 30\,s and a fixed initialisation
  transient occupies a substantially larger fraction of each sequence than
  of the longest evaluated clip (104.3\,s).

  \noindent On Session 1, DROID-W~\citep{droidw} and
  MASt3R-SLAM~\citep{mast3rslam} are substantially less robust and exhibit
  different failure modes. DROID-W suffers severe scale failure on 4 of 22
  episodes, with recovered scales of 0.80--2.68$\times$ and a mean SE(3)
  \ate{} of 395\,mm. On the remaining 18 episodes, it achieves a mean
  hand-eye \ate{} of 26.18\,mm at a mean fitted scale of 0.86$\times$.
  MASt3R-SLAM loses tracking on one episode and underestimates scale on all
  remaining 21 (0.23--0.66$\times$), yielding a mean hand-eye \ate{} of
  121.59\,mm. For \sysname, the mean scale deviation $|s-1|$ is 0.0120 on
  Session 1 and 0.0048 on Session 2, but fitted scale does not consistently
  predict episode-level accuracy ($r=+0.30$ and $r=-0.44$, respectively).

  \noindent On episodes of comparable duration (approximately 40\,s),
  \sysname requires an average of 206.0\,s per episode, compared with
  627.5\,s for DROID-W and 715.4\,s for MASt3R-SLAM. Under the same
  evaluation setting, these measurements correspond to approximately
  $3.0\times$ and $3.5\times$ reductions in per-episode wall-clock time,
  respectively.

\begin{table}[htbp]
\centering
\caption{
\textbf{Head-trajectory accuracy against \chingmu 26-camera 120\,Hz optical motion capture.}
Per-episode RMSE, averaged over episodes, hand-eye aligned per capture day (\autoref{tab:metrics}).
ORB-SLAM3 rows score an offline-refined trajectory, not the raw online output
(\autoref{sec:protocol}). \emph{Steady} discards the first 5\,s (not computed for
DROID-W/MASt3R-SLAM, Session 1 only). DROID-W/MASt3R-SLAM rows exclude failure episodes; see
\autoref{sec:accuracy}. \emph{Fitted scale} is not an error metric. \textbf{Bold}: lowest \ate{}
per group.
}
\label{tab:head-gt}
\begin{adjustbox}{max width=\linewidth}
\begin{tabular}{llcccccc}
\toprule
\textbf{Session} & \textbf{Head-pose source} & \textbf{n} & \textbf{Window} & \textbf{\ate RMSE} & \textbf{Rotation} & \textbf{RPE@1s} & \textbf{Fitted} \\
 & & & & mm & deg & mm / deg & scale \\
\midrule
Session 1 & ORB-SLAM3 \textit{(refined)} & 22 & full   & 3.27 & 0.184 & 3.77 / 0.219 & 0.9923 \\
     &                                       & 22 & steady & \textbf{2.12} & 0.154 & 2.64 / 0.188 & 0.9945 \\
     & DROID-W~\citep{droidw} \textit{(clean, excl.\ 4/22)} & 18 & full & 26.18 & --- & --- / --- & 0.8608 \\
     & MASt3R-SLAM~\citep{mast3rslam} \textit{(excl.\ 1/22)} & 21 & full & 121.59 & --- & --- / --- & 0.5268 \\
     & \sysname \textit{(ours, offline)}      & 22 & full   & 2.87 & 0.158 & 3.66 / 0.194 & 1.0118 \\
     &                                       & 22 & steady & 2.83 & 0.160 & 3.67 / 0.196 & 1.0117 \\
\midrule
Session 2 & ORB-SLAM3 \textit{(refined)} & 20 & full   & 1.00 & 0.260 & 1.34 / 0.308 & 1.0008 \\
     &                                       & 20 & steady & 0.92 & 0.263 & 1.29 / 0.317 & 1.0026 \\
     & \sysname \textit{(ours, offline)}      & 20 & full   & 0.75 & 0.258 & 1.06 / 0.307 & 0.9951 \\
     &                                       & 20 & steady & \textbf{0.65} & 0.262 & 1.04 / 0.316 & 0.9969 \\
\bottomrule
\end{tabular}
\end{adjustbox}
\end{table}

\subsection{Hand-Reconstruction Accuracy}
\label{sec:handacc}
\label{subsec:skeleton}
\label{subsec:hand-gt}
\label{subsec:hand-compare}
\label{subsec:headpose-source}

As shown in Table~\ref{tab:hand-compare}, \sysname outperforms all
open-source baselines on every metric. 

\noindent For \emph{detection}, \sysname attains perfect Precision, Recall, and F1 (1.00), producing neither
missed nor hallucinated hands across the entire evaluation set; in
contrast, Dyn-HaMR hallucinates frequently under occlusion (Precision
0.73 despite full recall), and HaWoR still misses or spuriously detects
hands occasionally (F1 0.98). Under the coverage-aware protocol, where
missed detections incur a deterministic placeholder error, complete
detection coverage also directly benefits the accuracy metrics below.

\noindent For \emph{3D pose}, \sysname reduces PA-MPJPE-p to 4.29\,mm, a 70\%
improvement over the strongest baseline HaWoR (16.30\,mm) and an
order-of-magnitude reduction relative to Dyn-HaMR (49.18\,mm).

\noindent The margin is largest on
\emph{orientation and position}: EPE-P drops from 83.10\,px (HaWoR) to
9.90\,px ($8.4\times$), and CT-p from 46.70 to 14.30 (69\% reduction),
indicating that our method recovers the absolute hand position far more
accurately, which we attribute to the explicit geometric depth
constraints provided by our multi-view depth estimation. 

\noindent For \emph{temporal smoothness}, \sysname attains a jitter of
1.11\,mm/frame$^2$, $3.2\times$ lower than the smoothest baseline
(Dyn-HaMR, 3.50) and an order of magnitude below HaWoR (16.71), without
any test-time optimization. Notably, HaPTIC fails entirely in our
multi-person capture scenes---a practically significant limitation, as
bystanders are common in real-world egocentric deployment. Overall,
these results demonstrate that \sysname delivers substantially more
complete, accurate, and stable hand reconstruction than existing
open-source methods.

\begin{table}[htbp]
\centering
\caption{
\textbf{Comparison with open-source egocentric hand reconstruction methods against
motion-capture ground truth.} All methods are evaluated on identical segments of
our motion-capture dataset.
All baselines are re-run and rescored on our data. HaPTIC fails to produce valid
output in our multi-person capture scenes. \textbf{Bold} marks the best
result in each column.
}
\label{tab:hand-compare}
\begin{adjustbox}{max width=\linewidth}
\begin{tabular}{lccccccc}
\hline
\multirow{2}{*}{\textbf{Method}} & \multicolumn{3}{c}{\textbf{Detection}} & \textbf{3D Pose} & \multicolumn{2}{c}{\textbf{Orient. \& Position}} & \textbf{Temporal} \\ \cmidrule(lr){2-4} \cmidrule(lr){5-5} \cmidrule(lr){6-7} \cmidrule(lr){8-8}
                                 & Precision\,$\uparrow$ & Recall\,$\uparrow$ & F1\,$\uparrow$ & PA-MPJPE-p\,$\downarrow$ & CT-p\,$\downarrow$ & EPE-P\,$\downarrow$ & Jitter\,$\downarrow$ \\ \hline
Dyn-HaMR\citep{dynhamr2025}      & 0.73           & 1.00       & 0.84     & 49.18            & 184.38                  & 96.42                  & 3.50              \\
HaWoR\citep{hawor2025}           & 0.96           & 0.99       & 0.98     & 16.30            & 46.70                   & 83.10                  & 16.71             \\
HaPTIC\citep{haptic2025}         & ---            & ---        & ---      & ---              & ---                     & ---                    & ---               \\ \hline
\textbf{\sysname (Ours)}         & \textbf{1.00}  & \textbf{1.00} & \textbf{1.00} & \textbf{4.29} & \textbf{14.30}      & \textbf{9.90}            & \textbf{1.11}     \\ \hline
\end{tabular}
\end{adjustbox}
\end{table}
\begin{figure*}[t]
\centering
\includegraphics[width=\linewidth]{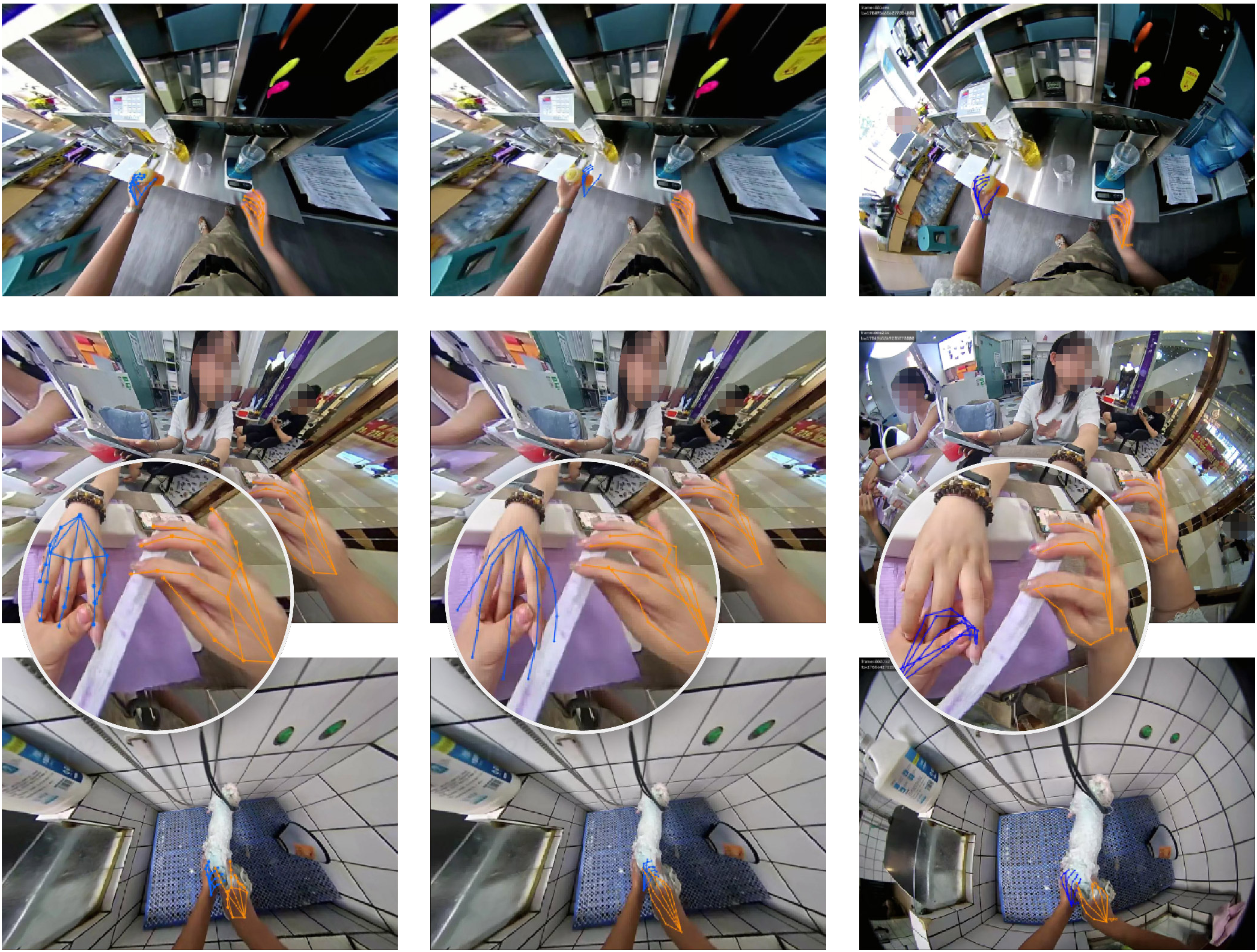}\\[3pt]
\begin{tabular}{@{}>{\centering\arraybackslash}p{\dimexpr0.315\linewidth\relax}@{\hspace{0.0261\linewidth}}%
>{\centering\arraybackslash}p{\dimexpr0.315\linewidth\relax}@{\hspace{0.0261\linewidth}}%
>{\centering\arraybackslash}p{\dimexpr0.315\linewidth\relax}@{}}
\textbf{Dyn-HaMR} & \textbf{HaWoR} & \textbf{\sysname (Ours)}
\end{tabular}
\caption{\textbf{Qualitative comparison on in-the-wild recordings}. \emph{Top}: rapid hand motion;
\emph{middle}: interaction with a bystander's hand;
\emph{bottom}: soiled hands under fast motion. Baselines exhibit pose offsets (top,
bottom) or misattribute the bystander's hand to the wearer (middle),
while \sysname maintains tight alignment and correct hand identity
throughout.}
\label{fig:wild}
\end{figure*}
\subsection{Qualitative results in the wild}
We further evaluate \sysname on real-world recordings for which no
motion-capture ground truth is available, and assess the results
qualitatively by the alignment between the projected hand skeletons and
the observed hands (Fig.~\ref{fig:wild}). Three representative
challenging scenarios are shown. 

\noindent (i)~\emph{Rapid hand motion}
(milk-tea preparation): both \sysname and Dyn-HaMR produce skeletons
that align well with the hands, whereas HaWoR exhibits a noticeable
offset on the left hand; this is consistent with its order-of-magnitude
higher jitter in Table~\ref{tab:hand-compare} (16.71 vs.\ 1.11), as
frame-wise instability manifests directly as misalignment under fast
motion. 

\noindent (ii)~\emph{Interaction with another person's hands} (manicure):
when the wearer's left hand approaches and interacts with a bystander's
left hand, both Dyn-HaMR and HaWoR misattribute the bystander's hand to
the wearer, while \sysname correctly disambiguates identities and
reconstructs only the wearer's hands. This failure mode explains the low
detection precision of Dyn-HaMR (0.73) in Table~\ref{tab:hand-compare}
and echoes the complete failure of HaPTIC in multi-person scenes:
distinguishing the wearer's hands from bystanders' hands remains an
open challenge for existing egocentric methods. 

\noindent (iii)~\emph{Rapid motion
with soiled hands} (animal washing): although both \sysname and Dyn-HaMR
successfully detect the dirt-covered hands, HaWoR shows a slight offset
on the left hand with an inaccurate wrist position, in line with its
higher CT-p error (46.70) in the quantitative comparison. 

\noindent Overall, the qualitative results corroborate our quantitative findings: \sysname
achieves more complete detection, tighter image-plane alignment, and
robust hand identity disambiguation, with the largest margins under fast
motion and multi-person interference.

\begingroup
\setlength{\emergencystretch}{3em}

\section{Scaling to Large-Scale Data Annotation}
\label{sec:scaling}

The original \sysname offline reconstruction pipeline achieves state-of-the-art metric-accurate 4D
head and bimanual hand reconstruction from wild multi-fisheye egocentric recordings. However, its
PyTorch research prototype suffers high runtime overhead and low GPU utilization, creating a
practical gap between lab-validated reconstruction and large-scale dataset production. Direct
execution of the unoptimized prototype cannot meet throughput requirements for generating
pseudo-supervision from thousands of real-world egocentric clips to pre-train embodied foundation
models.

\noindent To fill this gap, we introduce \depname, a production-grade end-to-end GPU engineering stack
built for \sysname. The original reconstruction formulation, metric constraints, anatomical loss
terms, and gravity-aligned world-frame outputs are fully preserved without any fidelity
degradation. \depname consists of four orthogonal engineering pillars
(\autoref{fig:deploy-pipeline}): hierarchical kernel acceleration, GPU memory and data-link
refinement, interleaved multi-job cluster scheduling, and automated speed-precision closed-loop
regression validation.

\subsection{Bottleneck Profiling of Vanilla \sysname}
\label{subsec:deploy-profiling}

Profiling the unmodified prototype exposes four key bottlenecks. The first is eager Python
overhead: fragmented PyTorch operators trigger frequent small kernel launches dominated by
Python-side logic. The second is heavy host-GPU traffic, as repeated CPU-GPU memory copies for
images, intermediate features and reconstruction states produce expensive memcpy overhead. Third,
the workflow is strictly sequential: detection, stereo depth inference, VIO alignment and bundle
adjustment run one after another, so GPU resources often stall waiting for CPU-bound decoding and
pre/post-processing. Finally, the prototype lacks production safeguards: unguided optimizations may
introduce metric-scale drift or temporal jitter, and no automatic regression check exists to catch
such quality drops.

\noindent All metric-critical components, including calibrated-stereo initialization, ego-aware
hand-ownership discrimination, MANO anatomical barriers and IMU gravity alignment, remain untouched.
Our optimizations only improve execution efficiency and do not alter reconstruction objectives.

\subsection{Hierarchical Kernel-Level Acceleration}
\label{subsec:deploy-kernel}

Multi-stage kernel optimizations deliver a 3.41$\times$ single-job speed-up for the core \sysname
forward pass (385 frames, including pre/post-processing; excluding file decoding, external detector
and full bundle adjustment). Hand detection, stereo depth and MANO regression networks are compiled
into FP16 TensorRT engines with operator fusion, whereas geometric solvers, scale-sensitive
constraints and anatomical regularization keep full FP32 precision to avoid kinematic artifacts.
Fisheye rectification, multi-view reprojection and MANO regularization are further re-implemented
as high-performance C++ hot-path kernels, and GPU computation overlaps with CPU I/O and
pre-processing to hide wall-clock latency. TensorRT mixed-precision compilation alone reduces the
baseline PyTorch latency from 10.191\,s to 7.203\,s, and adding the C++ kernels with compute
overlapping brings it down to 2.991\,s.

\subsection{GPU Memory and Data-Link Optimization}
\label{subsec:deploy-memory}

We redesign the end-to-end data flow to remove redundant cross-memory copies. A persistent GPU
buffer pool pre-allocates long-lived buffers for fisheye correction maps and intermediate
reconstruction variables, avoiding per-clip allocation-deallocation overhead. Static calibration
data---hardware intrinsics, extrinsics and rectification matrices---are uploaded to the GPU once and
reused for all clips captured by the identical \device rig. Host-device synchronization is kept to
a minimum: only compact final outputs (MANO parameters, 6-DoF head poses, world-space keypoints and
per-frame quality metrics) are copied back to the CPU, while large intermediate tensors stay
resident on the GPU.

\subsection{Interleaved Multi-Job Cluster-Aware Scheduling}
\label{subsec:deploy-scheduling}

Single-clip acceleration alone cannot saturate modern GPU hardware, so \depname adopts
cluster-oriented interleaved job scheduling. Compiled TensorRT engines persist in GPU memory as a
resident model pool, eliminating repeated model deserialization and CUDA-context reinitialization.
Whenever one clip is blocked by CPU decoding or I/O, the scheduler dispatches pending reconstruction
tasks to occupy idle GPU cycles. It also natively supports task-queue management, soft-failure
retry, multi-GPU workload partitioning and centralized quality-metric collection for dataset-scale
runs. With the complete stack, average GPU utilization rises from 34.1\% to 80.3\%, reducing idle
gaps between pipeline stages.

\subsection{Speed-Quality Closed-Loop Regression}
\label{subsec:deploy-regression}

Engineering optimizations can silently harm reconstruction quality, so every code change goes
through a four-step closed-loop validation. We first profile wall-clock latency, kernel timeline,
VRAM footprint and real GPU utilization via NVML, then attribute each bottleneck to computation
load, memory-copy overhead or CPU-side blocking, and then apply the corresponding remedy, such as
kernel fusion, buffer-pool tuning or scheduling adjustments. Finally, a regression test on held-out
motion-capture sequences audits head \ate RMSE, \pampjpe, scale deviation $|s-1|$, hand-detection
F1 and temporal jitter; any modification causing measurable quality degradation is rejected before
merging.

\subsection{End-to-End Large-Scale Evaluation}
\label{subsec:deploy-eval}

We benchmark CPU VIO plus the original serial handpose pipeline against the complete \depname
head-and-hand stack with ten worker slots on RTX 5090, including depth estimation and full bundle
adjustment.

\begin{table}[htbp]
\centering
\caption{
\textbf{Complete end-to-end configurations on the same nine clips.} Both use tmpfs output storage;
downloads and visualization are excluded. \textbf{Bold} marks the best result in each column.
}
\label{tab:deploy-perf}
\begin{adjustbox}{max width=\linewidth}
\begin{tabular}{lccccc}
\toprule
\textbf{Setup} & \textbf{Video} & \textbf{Wall-clock} & \textbf{Video : processing} & \textbf{Avg GPU} & \textbf{End-to-end} \\
 & duration (s) & time (s) & time & utilization & speed-up \\
\midrule
CPU VIO + original handpose \textit{(serial)} & 278.81 & 3094.1 & $1:11.097$ & 34.1\% & $1.00\times$ \\
\depname \textit{(10 worker slots)}           & 278.81 & \textbf{220.2} & $\mathbf{1:0.790}$ & \textbf{80.3\%} & $\mathbf{14.05\times}$ \\
\bottomrule
\end{tabular}
\end{adjustbox}
\end{table}

\begin{figure}[htbp]
\centering
\includegraphics[width=\linewidth]{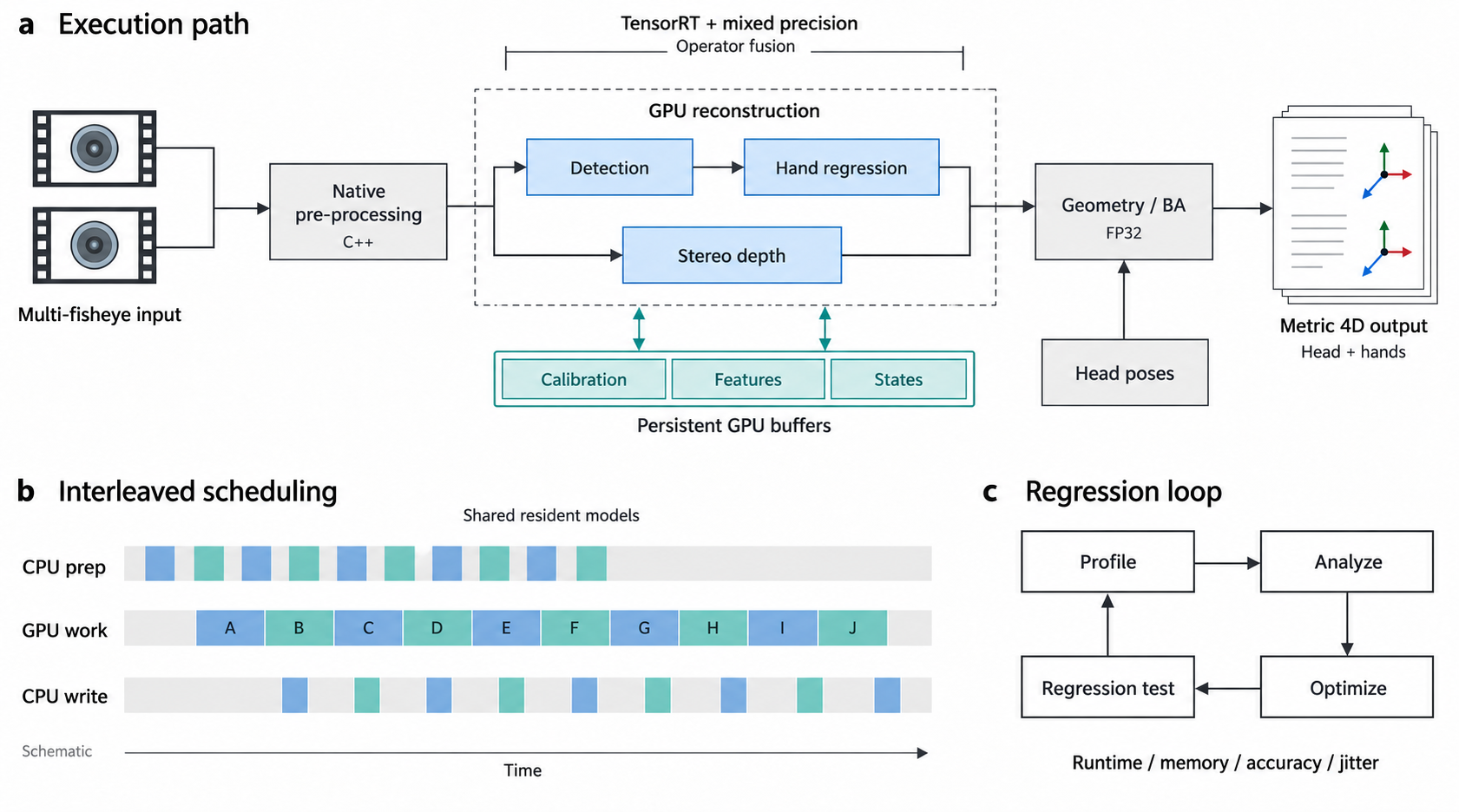}
\caption{
\textbf{The \sysname pipeline with \depname production branch.} Base \sysname
(\autoref{sec:method}) reconstructs metric 4D hand-head trajectories from multi-fisheye egocentric
input. The \depname branch (\autoref{sec:scaling}) is the production-oriented engineering stack for
large-scale dataset annotation, including kernel acceleration, memory-data-link refinement,
cluster-aware multi-job scheduling and automated speed-precision closed-loop regression. It yields
mass pseudo-labels with $14.05\times$ end-to-end throughput.
}
\label{fig:deploy-pipeline}
\end{figure}

\noindent As shown in \autoref{tab:deploy-perf}, total runtime is reduced by 92.9\%. All key
reconstruction metrics remain within regression tolerance, and hand-ownership detection stays
perfect ($\text{Precision}=\text{Recall}=\text{F1}=1.00$). Benefiting from this throughput boost, our
workflow can process tens-of-thousands of in-the-wild \device clips under cluster environments to
generate metric 4D hand-head pseudo-labels for embodied-model pre-training. Exported sequence-level
quality metrics enable automatic sample filtering without human inspection.

\subsection{Discussion and Limitations}
\label{subsec:deploy-limitations}

\depname unlocks industrial-scale automatic annotation while fully preserving \sysname's metric
reconstruction capability, but it comes with practical trade-offs. Its throughput gain is
scheduling-driven and requires sufficient queued jobs, so GPU utilization drops under light input
workloads. TensorRT and custom C++ kernels raise engineering maintenance cost, and network
architecture updates demand full regression cycles. The pipeline also remains offline, targeting
dataset generation rather than low-latency real-time streaming inference. Finally, cluster-scale
annotation demands abundant GPU resources: single-GPU execution still achieves kernel-level
acceleration but cannot reach peak hardware utilization.

\noindent In summary, \depname bridges lab-grade reconstruction and production-ready large-scale
pseudo-annotation. It enables \sysname's high-fidelity metric 4D reconstruction to scale toward
massive real-world egocentric datasets for embodied manipulation foundation-model training.

\par\endgroup

\section{Conclusion}
\label{sec:conclusion}

\sysname{} takes an unprepared egocentric recording and returns metric two-hand motion and a metric
head trajectory in one world frame, and --- unusually for this class of system --- it has been held to
a measured standard: both delivered quantities were scored against 26-camera optical motion capture
under a protocol that charges missed detections and audits its own reference rather than assuming it.
Head trajectory lands at 0.65 to 2.83\,mm, finger articulation near 5\,mm \pampjpe --- the last being the figure a flattering summary
would omit and a consumer of the labels most needs. What transfers beyond this particular system is
narrower than the system itself: taking metric gauge from calibration at initialisation, rather than
recovering it by alignment afterwards, is what makes such a measurement meaningful at all, and the
quantity that exposes its failure is fitted scale rather than aligned error. Whether the next
millimetre requires a better estimator or a better reference is genuinely undecided, and the board
capture of \autoref{subsec:board} is the experiment that would decide it.

\section{Contributions and Acknowledgments}
\label{sec:contributions}

% Layout mirrors the reference report: bold role labels flush left on their own
% line, names on a 94pt column pitch (first column flush at the margin), and a
% flush-left acknowledgments paragraph.
\newlength{\ncw}\setlength{\ncw}{94pt}
\newcommand{\namecols}[1]{%
  \par\noindent\begingroup
  \setlength{\tabcolsep}{0pt}%
  \renewcommand{\arraystretch}{0}%
  \begin{tabular}{@{}*{5}{>{\raggedright\arraybackslash}p{\ncw}}@{}}%
  #1\end{tabular}%
  \endgroup\par}

\newcommand{\rolelabel}[1]{\par\noindent\textbf{#1}\par}

\vspace{7.1pt}
\rolelabel{Contributors}
\vspace{3.7pt}
\namecols{%
Jiangong Xiao & Chao Ma & Zhihao Zhang & Yifei Dong & Zhouyi Jin \\[3.65pt]
Zhiwen Hou & Haoyi Wang & Junle Yu & Amu Zheng & Chen Wang \\[3.65pt]
Kaishu Lin & Hanwen Shi & Shiyun Zhang & Renquan Wan & Jingmei Jin \\[3.65pt] 
Xuan Hu & Maoqing Li & Li Liu & Maoqing Yao}

\vspace{6.7pt}
\rolelabel{Academic Advisors}
\vspace{-0.4pt}
\namecols{Weihuang Chen & Hongbin Sun & Jinli Suo}

\vspace{4.7pt}
\rolelabel{Project Leads}
\vspace{-1.7pt}
\namecols{Li Liu & Maoqing Li}

\vspace{6.7pt}
\rolelabel{Corresponding Author}
\vspace{-0.2pt}
\namecols{Maoqing Yao}

\vspace{4.0pt}
\rolelabel{Acknowledgments}
\vspace{-3.7pt}
\noindent We would like to sincerely thank the broader team for their
tremendous support, including Jia Zeng, Yu'an Zhang, Zhitan Zheng, Jun Ji, Fu Fang, Yuwei Jia, Pengyong Wang, Zheyuan Wang, Wei Yu, Xiaofan Zhang and Yongjian Shen for their valuable discussions,
review, motion-capture room setup, and computing and project support.

\bibliographystyle{unsrtnat}
\bibliography{main}

\end{document}